\documentclass[11pt,letterpaper]{article}

\usepackage[T1]{fontenc}
\usepackage[utf8]{inputenc}
\usepackage{lmodern}
\usepackage[margin=1in]{geometry}
\usepackage{microtype}
\usepackage{amsmath,amssymb,bm}
\usepackage{graphicx,booktabs}
\usepackage[accsupp]{axessibility}
\usepackage{xcolor}
\usepackage{tikz}
\usetikzlibrary{arrows.meta,positioning,calc,fit,backgrounds,angles,quotes}
\usepackage[font=small,labelfont=bf]{caption}

\usepackage[numbers,sort&compress]{natbib}
\usepackage{xurl}
\usepackage[hidelinks]{hyperref}
\usepackage{orcidlink}
\hypersetup{
    pdftitle={LoRA Superposition across Pretrained Models},
    pdfauthor={Keigo Watanabe, Hiroshi Kera, Kazuhiko Kawamoto},
    pdfsubject={LoRA superposition across pretrained models},
    pdfkeywords={low-rank adaptation, LoRA, parameter superposition, model merging, vision transformer, pretrained models}
}

\title{\LARGE\bfseries LoRA Superposition across Pretrained Models}
\author{%
    Keigo Watanabe\textsuperscript{1}\,\orcidlink{0009-0008-3755-234X}%
    \thanks{The first and third authors contributed equally to this work and share first authorship.}
    \quad Hiroshi Kera\textsuperscript{1,2}
    \quad Kazuhiko Kawamoto\textsuperscript{1}\,\orcidlink{0000-0003-3701-1961}\footnotemark[1]\\[0.6em]
    \small\textsuperscript{1}Chiba University, Chiba, Japan\\
    \small\textsuperscript{2}National Institute of Informatics, Tokyo, Japan\\[0.3em]
    \small\texttt{\{keigo.watanabe,kera\}@chiba-u.jp}\quad
    \texttt{kawa@faculty.chiba-u.jp}%
}
\date{}

\begin{document}
\maketitle

\begin{abstract}
Low-rank adaptation (LoRA) is usually tied to the pretrained backbone on which it is learned. We ask whether LoRAs trained independently for the same task on different pretrained backbones can be summed and applied unchanged to all contributing backbones. We call this construction \emph{LoRA superposition}. Across seven ViT-B/16 CLIP backbones and eight image-classification tasks, the exact sum improves all 56 backbone--task pairs and achieves a mean accuracy gain of 30.21 percentage points over the unadapted backbone. The exact sum achieves a mean accuracy only 0.28 percentage points below that of individually trained solo LoRAs. In contrast, a leave-one-model-out (LOMO) sum produces a mean gain of only 0.06 percentage points over the unadapted backbones. Backbone-specific solo LoRAs are nearly orthogonal in parameter space, while pairwise cross-backbone responses are strongly concentrated on the diagonal. These results support a component-wise non-interference hypothesis, under which the frozen backbone acts as an implicit selector of its own component without explicit routing or cross-model alignment.
\end{abstract}

\section{Introduction}
\label{sec:intro}

Low-rank adaptation (LoRA)~\cite{hu2022lora} adapts a pretrained model
through low-rank parameter updates while keeping the original backbone
fixed.
A LoRA update can also be viewed as a task vector
~\cite{ilharco2023task} and reused through parameter-space operations.
However, a LoRA is typically tied to the backbone on which it was
trained.
Even for backbones with the same architecture, different pretrained
weights generally require separate adapters.
Supporting $T$ tasks with $M$ pretrained backbones therefore usually
requires $M\times T$ adapters, limiting the reuse of LoRA libraries
across pretrained models.

Existing approaches address related forms of parameter reuse.
Task-vector merging~\cite{yadav2023ties,yu2024dare} and
adapter-library methods~\cite{huang2024lorahub,ostapenko2024arrow}
typically combine updates on a common pretrained backbone.
Cross-model transfer instead maps updates between backbones through
orthogonal alignment~\cite{kawamoto2025orthogonal}, parameter
re-basing~\cite{rinaldi2025transfusion}, or shared subspaces
~\cite{farhadzadeh2025lorax}.
Model-superposition methods store multiple models or tasks in shared
parameters, but require contexts, masks, or explicit selection
mechanisms~\cite{cheung2019superposition,wortsman2020supermasks,
zeman2023superformer}.
It therefore remains unclear whether backbone-specific LoRAs can
coexist directly in one update without alignment or explicit routing.

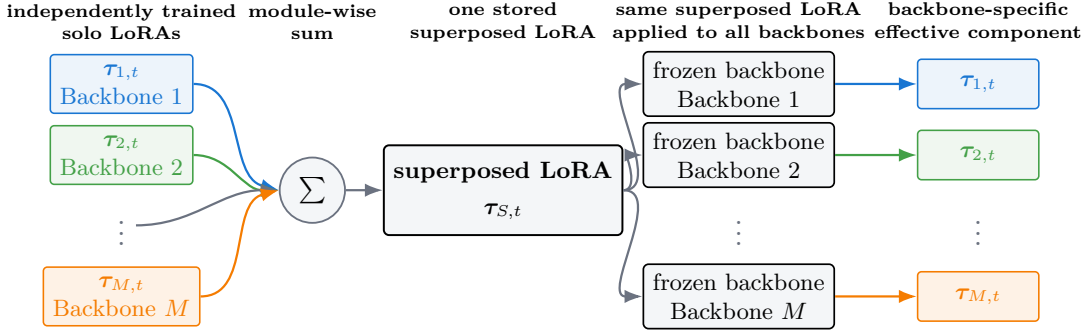
\begin{figure}[!t]
    \centering
    \resizebox{0.9\textwidth}{!}{
        \begin{tikzpicture}[
    >=Latex,
    font=\fontsize{9}{11}\selectfont,
    solo/.style={
        draw,
        rounded corners=1.5pt,
        minimum width=1.55cm,
        minimum height=0.66cm,
        align=center,
        line width=0.6pt
    },
    backbone/.style={
        draw,
        rounded corners=2pt,
        minimum width=1.90cm,
        minimum height=0.82cm,
        align=center,
        line width=0.65pt,
        fill=cLight
    },
    response/.style={
        draw,
        rounded corners=1.5pt,
        minimum width=1.62cm,
        minimum height=0.66cm,
        align=center,
        line width=0.6pt
    },
    flow/.style={
        ->,
        line width=0.85pt
    },
    stage/.style={
        font=\fontsize{7}{8.4}\selectfont\bfseries,
        align=center
    }
]

\definecolor{cOne}{HTML}{1976D2}
\definecolor{cTwo}{HTML}{43A047}
\definecolor{cM}{HTML}{F57C00}
\definecolor{cGray}{HTML}{6B7280}
\definecolor{cLight}{HTML}{F4F6F8}

\node[stage] at (0,2.25)
{independently trained\\solo LoRAs};

\node[
    solo,
    draw=cOne,
    fill=cOne!8,
    text=cOne
] (t1) at (0,1.42)
{$\bm{\tau}_{1,t}$\\Backbone 1};

\node[
    solo,
    draw=cTwo,
    fill=cTwo!8,
    text=cTwo
] (t2) at (0,0.47)
{$\bm{\tau}_{2,t}$\\Backbone 2};

\node[
    font=\large,
    text=cGray
] (tdots) at (0,-0.47)
{$\vdots$};

\node[
    solo,
    draw=cM,
    fill=cM!10,
    text=cM
] (tM) at (0,-1.42)
{$\bm{\tau}_{M,t}$\\Backbone $M$};

\node[stage] at (2.55,2.25)
{module-wise\\sum};

\node[
    circle,
    draw=cGray,
    fill=cLight,
    minimum size=0.76cm,
    line width=0.7pt
] (sum) at (2.55,0)
{$\sum$};

\draw[flow,cOne]
    (t1.east) to[out=0,in=165] (sum.west);

\draw[flow,cTwo]
    (t2.east) to[out=0,in=180] (sum.west);

\draw[flow,cGray]
    (tdots.east) to[out=0,in=180] (sum.west);

\draw[flow,cM]
    (tM.east) to[out=0,in=195] (sum.west);

\node[stage] at (5.10,2.25)
{one stored\\superposed LoRA};

\node[
    draw,
    rounded corners=2pt,
    minimum width=2.10cm,
    minimum height=1.20cm,
    align=center,
    line width=0.75pt,
    fill=cLight
] (sup) at (5.10,0)
{
    \textbf{superposed LoRA}\\[0.3em]
    $\bm{\tau}_{S,t}$
};

\draw[flow,cGray]
    (sum.east) -- (sup.west);

\node[stage] at (8.25,2.25)
{same superposed LoRA\\applied to all backbones};

\node[backbone] (m1) at (8.25,1.42)
{frozen backbone\\Backbone 1};

\node[backbone] (m2) at (8.25,0.47)
{frozen backbone\\Backbone 2};

\node[
    font=\large,
    text=cGray
] (mdots) at (8.25,-0.47)
{$\vdots$};

\node[backbone] (mM) at (8.25,-1.42)
{frozen backbone\\Backbone $M$};

\draw[flow,cGray]
    (sup.east) to[out=0,in=180] (m1.west);

\draw[flow,cGray]
    (sup.east) to[out=0,in=180] (m2.west);

\draw[flow,cGray]
    (sup.east) to[out=0,in=180] (mM.west);

\node[stage] at (11.45,2.25)
{backbone-specific\\effective component};

\node[
    response,
    draw=cOne,
    fill=cOne!8,
    text=cOne
] (r1) at (11.45,1.42)
{$\bm{\tau}_{1,t}$};

\node[
    response,
    draw=cTwo,
    fill=cTwo!8,
    text=cTwo
] (r2) at (11.45,0.47)
{$\bm{\tau}_{2,t}$};

\node[
    font=\large,
    text=cGray
] (rdots) at (11.45,-0.47)
{$\vdots$};

\node[
    response,
    draw=cM,
    fill=cM!10,
    text=cM
] (rM) at (11.45,-1.42)
{$\bm{\tau}_{M,t}$};

\draw[flow,cOne]
    (m1.east) -- (r1.west);

\draw[flow,cTwo]
    (m2.east) -- (r2.west);

\draw[flow,cM]
    (mM.east) -- (rM.west);

\end{tikzpicture}
    }
\caption{
Cross-backbone LoRA superposition.
Backbone-specific solo LoRAs for the same task are summed into one
shared adapter, which is stored and applied unchanged to all
contributing frozen backbones.
Under the non-interference hypothesis, each backbone responds mainly to its own component in the superposed update.
The exact sum is a composite update whose rank can grow with the number of contributing backbones.
}
   \label{fig:concept}
\end{figure}

We study LoRAs trained independently for the same task on backbones
with the same architecture but different pretrained weights.
We refer to each backbone-specific LoRA as a \emph{solo LoRA} and
define \emph{LoRA superposition} by simply summing their parameter
updates.
The same summed LoRA is then applied unchanged to every contributing
backbone.
For each task, only the summed update is saved and distributed as one
shared adapter.
No additional optimization, calibration data, model identifier, or
target-specific transformation is required after the solo LoRAs have
been trained.
Figure~\ref{fig:concept} illustrates the construction.
Our central hypothesis is component-wise non-interference.
Backbone-specific solo LoRAs occupy nearly orthogonal directions in
parameter space, while each frozen backbone responds primarily to its
own component.
The frozen backbone can therefore act as an implicit selector within
the superposed update.
This hypothesis predicts that the full sum should retain the
performance of the target backbone's solo LoRA, whereas a
leave-one-model-out (LOMO) sum that excludes its own component should
have little effect.

We evaluate LoRA superposition on seven ViT-B/16 CLIP backbones and
eight image-classification tasks.
The exact sum improves all 56 backbone--task pairs
and achieves a mean accuracy gain of 30.21 percentage points.
Its accuracy remains close to Solo, whereas LOMO sum improves
over Backbone by only 0.06 percentage points on average.
Parameter-space similarity and pairwise cross-backbone response
further support the component-wise non-interference hypothesis.

Our contributions are twofold.
\begin{itemize}
    \item
    We formulate LoRA superposition, which sums same-task,
    backbone-specific solo LoRAs and applies the resulting update
    unchanged to all contributing backbones without alignment or
    explicit routing.

    \item
    We characterize component-wise non-interference through
    parameter-space similarity, pairwise cross-backbone response,
    and target-excluded LOMO analysis across seven backbones and
    eight tasks.
\end{itemize}

\section{Related Work}

\subsection{Task Vectors and LoRA Composition}

Task arithmetic~\cite{ilharco2023task} represents a fine-tuning
displacement as a reusable direction in parameter space and shows
that task-specific updates can be combined through simple arithmetic
operations.
TIES-Merging~\cite{yadav2023ties} and
DARE~\cite{yu2024dare} address interference when combining multiple
task vectors.
Task Singular Vectors~\cite{gargiulo2025tsv} further exploit
low-dimensional structure in task updates to reduce interference
during model merging.
For LoRA adapters, LoRAHub~\cite{huang2024lorahub} and
Arrow~\cite{ostapenko2024arrow} select or compose adapters from a
library according to the target task.
These approaches demonstrate that parameter updates can be reused and
combined effectively, but they generally assume that the updates are
defined relative to a common pretrained backbone.
Our setting instead considers same-task updates learned independently
from different pretrained backbones.

\subsection{Cross-Model Adaptation}

Cross-model task-vector methods aim to make an update learned on one
pretrained model usable on another.
Existing approaches transport updates between models through subspace
projection~\cite{farhadzadeh2025lorax}, parameter
re-basing~\cite{rinaldi2025transfusion}, or learned orthogonal
alignment~\cite{kawamoto2025orthogonal}.
These methods address the mismatch between the parameter spaces or
representations of independently pretrained models.
However, cross-model adaptation generally requires a model-pair
correspondence, target-side subspace, or calibration data.
The resulting transformation is therefore constructed for a
particular source--target relationship.
LoRA superposition considers a different setting: no update is
transformed for another backbone, and the same summed update is
applied unchanged to all contributing backbones.

\subsection{Model and Parameter Superposition}

Parameter superposition
(PSP)~\cite{cheung2019superposition} stores multiple task-specific
networks in shared parameters and uses context vectors to recover the
desired network.
Supermasks in Superposition
(SupSup)~\cite{wortsman2020supermasks} represents each task with a
binary mask over a fixed network and selects or infers the active mask
at test time.
SuperFormer~\cite{zeman2023superformer} extends context-based
superposition to Transformer models for continual learning.
These methods show that multiple parameter configurations can coexist
within a shared representation.
However, existing model-superposition methods require a context,
mask, or task-selection mechanism to retrieve the desired component.
In our setting, no explicit selector is introduced; instead, the
frozen backbone itself is hypothesized to respond selectively to its
own component in the summed LoRA.

\subsection{LoRA Merging}

LoRA-LEGO~\cite{zhao2025loralego} merges LoRAs through rank-wise
clustering, while KnOTS~\cite{stoica2025knots} aligns LoRA updates
before applying model-merging methods.
These methods address interference or incompatibility among multiple
LoRA updates and generally consider LoRAs defined for a common
backbone.
In contrast, our setting sums same-task LoRAs trained independently
for different pretrained backbones.

LoRA superposition therefore differs from task-vector merging,
cross-model transfer, conventional parameter superposition, and LoRA
merging.
The same summed update is applied unchanged to all contributing
backbones without parameter alignment, context vectors, or explicit
routing.
Our focus is whether the backbone-specific effects remain functionally
separable after this direct addition.

\section{LoRA Superposition}

\subsection{Problem Formulation}

Consider a fixed downstream task $t$ and a closed set
$S\subseteq\{1,\ldots,M\}$ of $K=|S|$ pretrained backbones.
The backbones have the same architecture and the same ordered set of
$J$ corresponding weight matrices.
Their pretrained weights may differ.
All backbone parameters remain frozen.
A separate solo LoRA is trained for each backbone in $S$ on task $t$.

For backbone $m$ and module $j$, a rank-$r$ solo LoRA produces
\begin{equation}
    \bm{W}_{m,j}'
    =
    \bm{W}_{m,j}
    +
    \Delta\bm{W}_{m,t,j},
    \quad
    \Delta\bm{W}_{m,t,j}
    =
    \bm{B}_{m,t,j}\bm{A}_{m,t,j}.
    \label{eq:solo-update}
\end{equation}
The LoRA scaling factor is absorbed into $\bm{B}_{m,t,j}$.

We represent the complete solo LoRA as a vector obtained by
vectorizing and concatenating all module updates:
\begin{equation}
    \bm{\tau}_{m,t}
    =
    \begin{bmatrix}
        \operatorname{vec}(\Delta\bm{W}_{m,t,1})\\
        \vdots\\
        \operatorname{vec}(\Delta\bm{W}_{m,t,J})
    \end{bmatrix}.
    \label{eq:solo-lora}
\end{equation}
We use the effective updates
$\Delta\bm{W}_{m,t,j}=\bm{B}_{m,t,j}\bm{A}_{m,t,j}$
rather than the LoRA factors themselves.
This avoids the non-uniqueness of the LoRA factorization.

LoRA superposition is defined as the sum of the solo LoRA vectors:
\begin{equation}
    \bm{\tau}_{S,t}
    =
    \sum_{m\in S}\bm{\tau}_{m,t}.
    \label{eq:superposition}
\end{equation}
Equivalently, each corresponding module update is summed as
\begin{equation}
    \Delta\bm{W}_{S,t,j}
    =
    \sum_{m\in S}\Delta\bm{W}_{m,t,j}.
    \label{eq:module-superposition}
\end{equation}
For deployment, we store only the summed module matrices
$\{\Delta\bm{W}_{S,t,j}\}_{j=1}^{J}$ as one shared adapter.
The individual solo LoRAs are not needed after the sum is formed.
Each saved matrix has the same dimensions as the corresponding backbone
weight, independent of $K$.
Because each component has rank at most $r$, the summed matrix has rank
at most $Kr$, or 112 in our setting with $K=7$ and $r=16$.
The single stored adapter can therefore have higher-rank updates than a
solo LoRA.

The same superposed LoRA is applied unchanged to every contributing
backbone:
\begin{equation}
    \bm{W}_{i,j}^{\mathrm{sup}}
    =
    \bm{W}_{i,j}
    +
    \Delta\bm{W}_{S,t,j},
    \quad i\in S.
    \label{eq:application}
\end{equation}
After the solo LoRAs are trained, the construction requires no
additional data, optimization, model identifier, or target-specific
transformation.
A same-architecture backbone $i\notin S$ serves as a target-excluded
control.

\begin{figure}[!t]
\centering
\resizebox{0.7\columnwidth}{!}{%
\begin{tikzpicture}[
    x={(0.95cm,0cm)},
    y={(-0.75cm,-0.42cm)},
    z={(0cm,1.0cm)},
    >=Latex,
    font=\fontsize{10}{12}\selectfont,
    vecA/.style={->, very thick, blue},
    vecB/.style={->, very thick, orange},
    vecC/.style={->, very thick, teal!70!black},
    vecS/.style={->, very thick, violet},
    proj/.style={dashed, thick, gray},
    axis/.style={->, semithick}
]

\coordinate (O) at (0,0,0);
\coordinate (X) at (3.5,0,0);
\coordinate (Y) at (0,3.0,0);
\coordinate (Z) at (0,0,3.5);

\coordinate (A) at (1.2,0,1.2);
\coordinate (B) at (-0.9,1.8,0.9);
\coordinate (C) at (-1.1,-1.1,1.1);

\coordinate (S) at (-0.8,0.7,3.2);

\draw[axis] (O) -- (X);
\draw[axis] (O) -- (Y);
\draw[axis] (O) -- (Z);

\node[below left, xshift=8pt] at (O) {$O$};

\draw[vecA]
    (O) -- (A)
    node[pos=0.62, left=3pt,xshift=28pt,yshift=-5pt]
    {$\bm{\tau}_{A,t}$};

\draw[vecB]
    (O) -- (B)
    node[pos=0.62, left=3pt,yshift=-10pt]
    {$\bm{\tau}_{B,t}$};

\draw[vecC]
    (O) -- (C)
    node[pos=0.55, left=3pt,xshift=8pt, yshift=25pt]
    {$\bm{\tau}_{C,t}$};

\draw[vecS]
    (O) -- (S)
    node[
        pos=0.62,
        left=4pt,
        xshift=25pt,
        yshift=40pt
    ]
    {$\bm{\tau}_{S,t}
      =\bm{\tau}_{A,t}+\bm{\tau}_{B,t}+\bm{\tau}_{C,t}$};

\draw[proj] (S) -- (A);

\pic[
    draw=gray,
    line width=0.7pt,
    angle radius=7pt
] {right angle = O--A--S};

\fill[blue] (A) circle (1.8pt);

\node[
    anchor=west,
    yshift=8pt,
    xshift=-3pt
] at (A)
{$\bm{\Pi}_{A}\bm{\tau}_{S,t}
\approx\bm{\tau}_{A,t}$};

\end{tikzpicture}
}
\caption{
Schematic geometry of LoRA superposition.
For backbone $A$, the superposed update is projected onto
$\bm{\tau}_{A,t}$, giving
$\bm{\Pi}_A\bm{\tau}_{S,t}\approx\bm{\tau}_{A,t}$.
}
\label{fig:projection-geometry}
\end{figure}
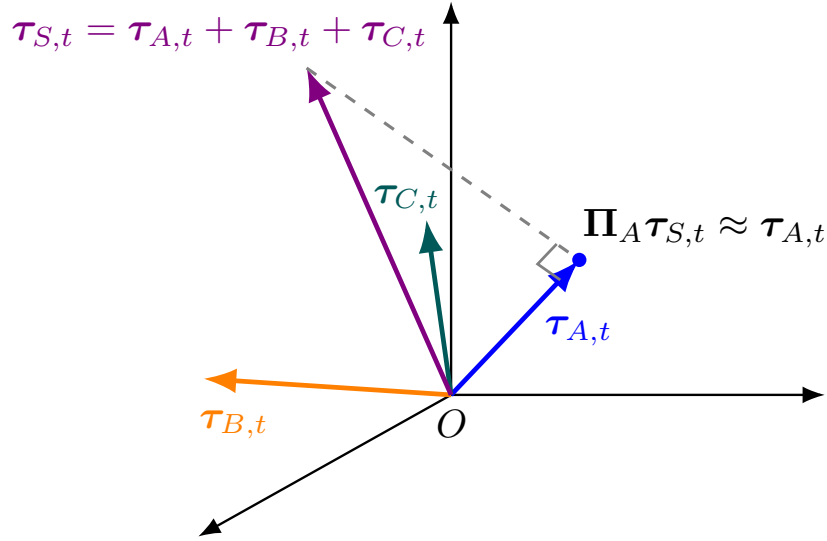

\subsection{Non-Interference Hypothesis}

Let $a_{i,t}(\bm{\tau})$ denote the accuracy of backbone $i$ on task $t$
after applying LoRA $\bm{\tau}$.
The notation $a_{i,t}(\bm{0})$ denotes the accuracy without a LoRA update.

Our hypothesis is motivated by the geometry of the backbone-specific
LoRA updates.
Solo LoRAs trained for different backbones are nearly orthogonal in
parameter space.
At the same time, each frozen backbone is sensitive only to a limited
set of update directions that are useful for that backbone.
We therefore interpret backbone $i$ as inducing an implicit projection
onto its effective update directions.
Figure~\ref{fig:projection-geometry} schematically illustrates this
interpretation: nearly orthogonal solo LoRAs coexist in the
superposed update, while backbone $i$ effectively retains its own
component.
If $\bm{\Pi}_i$ denotes this conceptual projection, the expected behavior is
\begin{equation}
    \bm{\Pi}_i\bm{\tau}_{i,t}\approx\bm{\tau}_{i,t},
    \quad
    \bm{\Pi}_i\bm{\tau}_{m,t}\approx\bm{0}
    \quad (m\neq i).
    \label{eq:implicit-projection}
\end{equation}
No projection is explicitly computed in our method.
Equation~\eqref{eq:implicit-projection} only describes how the frozen
backbone responds to different update directions.

For the superposed LoRA
$\bm{\tau}_{S,t}=\sum_{m\in S}\bm{\tau}_{m,t}$, this interpretation gives
\begin{equation}
    \bm{\Pi}_i\bm{\tau}_{S,t}
    =
    \sum_{m\in S}\bm{\Pi}_i\bm{\tau}_{m,t}
    \approx
    \bm{\tau}_{i,t},
    \quad i\in S.
    \label{eq:implicit-selection}
\end{equation}
Thus, multiple nearly orthogonal LoRA components can coexist in the
same update, while each backbone effectively selects the component
aligned with its own useful directions.

This geometric interpretation leads to two empirical predictions.
First, backbone $i$ should respond only weakly to a solo LoRA trained
for another backbone:
\begin{equation}
    a_{i,t}(\bm{\tau}_{m,t})
    \approx
    a_{i,t}(\bm{0}),
    \quad m\neq i.
    \label{eq:pairwise-noninterference}
\end{equation}
Second, the superposed LoRA should preserve the accuracy obtained by
the target backbone's own solo LoRA:
\begin{equation}
    a_{i,t}(\bm{\tau}_{S,t})
    \approx
    \begin{cases}
        a_{i,t}(\bm{\tau}_{i,t}), & i\in S,\\
        a_{i,t}(\bm{0}), & i\notin S.
    \end{cases}
    \label{eq:noninterference}
\end{equation}

For an included backbone $i$, define the other-backbone sum as
\begin{equation}
    \bm{\tau}_{S\setminus\{i\},t}
    =
    \sum_{m\in S\setminus\{i\}}\bm{\tau}_{m,t}.
    \label{eq:other-components}
\end{equation}
The hypothesis therefore predicts
\begin{equation}
\begin{aligned}
    a_{i,t}
    \left(
    \bm{\tau}_{i,t}+\bm{\tau}_{S\setminus\{i\},t}
    \right)
    &\approx a_{i,t}(\bm{\tau}_{i,t}),\\
    a_{i,t}
    \left(
    \bm{\tau}_{S\setminus\{i\},t}
    \right)
    &\approx a_{i,t}(\bm{0}).
\end{aligned}
    \label{eq:component-predictions}
\end{equation}
A model-invariant direction should remain effective after the target
backbone's own component is removed.
The target-excluded case therefore distinguishes such a direction
from closed-set superposition.

\section{Experiments}

\subsection{Setup}

\begin{figure}[!t]
    \centering
    \includegraphics[
        width=\columnwidth
    ]{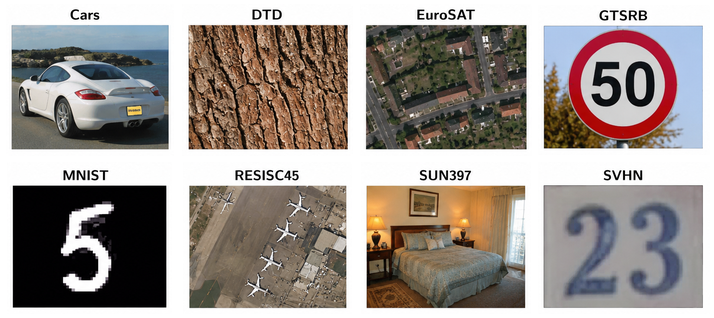}
    \caption{
    Representative samples from the eight downstream datasets.}
    \label{fig:dataset-samples}
\end{figure}

\noindent\textbf{Datasets.}
We evaluate eight image-classification datasets:
Cars~\cite{krause2013cars},
DTD~\cite{cimpoi2014dtd},
EuroSAT~\cite{helber2019eurosat},
GTSRB~\cite{stallkamp2011gtsrb},
MNIST~\cite{lecun1998mnist},
RESISC45~\cite{cheng2017resisc45},
SUN397~\cite{xiao2010sun397},
and SVHN~\cite{netzer2011svhn}.
Representative samples are shown in
Fig.~\ref{fig:dataset-samples}.
We follow the dataset splits used in
task arithmetic~\cite{ilharco2023task}.
Validation data are not used for hyperparameter tuning or early stopping.

\noindent\textbf{Backbones.}
We use seven publicly available ViT-B/16 CLIP backbones distributed
through OpenCLIP~\cite{radford2021clip,ilharco2021openclip}:
OpenAI, LAION-400M, LAION-2B,
DataComp-1B~\cite{gadre2023datacomp},
DFN-2B~\cite{fang2024dfn},
MetaCLIP-400M, and MetaCLIP-2.5B~\cite{xu2024metaclip}.
All backbones share the same vision-transformer architecture but
differ in their pretraining data and curation strategies, as
summarized in Table~\ref{tab:pretraining-data}.
For each backbone, we construct fixed task-specific classifiers
by prompt ensembling with its pretrained text
encoder~\cite{radford2021clip,ilharco2023task}.
We use the same standard OpenCLIP training and evaluation
transforms for all backbones.

\noindent\textbf{Training.}
For both Solo and Joint shared, LoRA is applied only to the
$\bm{Q}$ and $\bm{V}$ projections in all 12 attention blocks, giving
$J=24$ modules, with rank $r=16$ for each module.
We use LoRA scaling $\alpha/r$ with $\alpha=32$.
All backbone parameters remain frozen.
We minimize the cross-entropy loss.
The number of fine-tuning epochs is dataset-dependent and is identical for Solo and Joint shared: 35 for Cars, 76 for DTD, 12 for EuroSAT, 11 for GTSRB, 5 for MNIST, 15 for RESISC45, 14 for SUN397, and 4 for SVHN.
For Solo, a separate LoRA is trained for each backbone–task pair. For Joint shared, the same mini-batch is evaluated by all seven backbones at each step. The seven losses are averaged with equal weights, and the common LoRA is updated once.

\begin{table}[tbp]
\centering
\small
\caption{
Pretraining data of the seven CLIP backbones.
All datasets contain web-scale image--text pairs but differ in
source pools, filtering criteria, and dataset sizes.
}
\label{tab:pretraining-data}
\begin{tabular}{lcll}
\toprule
Backbone & Data scale & Source & Main curation strategy\\
\midrule
OpenAI
& 400M
& Public Web data
& Proprietary collection and filtering\\

LAION-400M
& 400M
& Common Crawl
& CLIP-score filtering\\

LAION-2B
& 2.3B
& Common Crawl
& CLIP-score and language filtering\\

DataComp-1B
& 1.4B
& CommonPool
& Image-based and CLIP-score filtering\\

DFN-2B
& 2B
& CommonPool
& Learned data-filtering network\\

MetaCLIP-400M
& 400M
& Common Crawl
& Metadata-based concept balancing\\

MetaCLIP-2.5B
& 2.5B
& Common Crawl
& Metadata-based concept balancing\\
\bottomrule
\end{tabular}
\end{table}

\noindent\textbf{Compared methods.}
We compare the proposed \emph{Exact sum} with three reference
settings---\emph{Backbone}, \emph{Solo}, and \emph{Joint shared}---
and a target-excluded diagnostic, \emph{LOMO sum}.
\begin{itemize}
    \setlength{\itemsep}{0.15em}
    \setlength{\topsep}{0.25em}

    \item
    \emph{Backbone} applies no LoRA update and provides the
    unadapted reference.

    \item
    \emph{Solo} applies the backbone-specific LoRA
    $\bm{\tau}_{i,t}$.

    \item
    \emph{Joint shared} jointly trains one common rank-16 LoRA
    across all seven frozen backbones for each task with equal
    weighting.

    \item
    \emph{Exact sum} applies the post-hoc superposition
    $\bm{\tau}_{S,t}$ defined in Eq.~\eqref{eq:superposition}.

    \item
    \emph{LOMO sum} applies the target-excluded superposition
    \begin{equation}
        \bm{\tau}_{S\setminus\{i\},t}
        =
        \sum_{m\in S\setminus\{i\}}
        \bm{\tau}_{m,t}
        \label{eq:lomo-sum}
    \end{equation}
    to target backbone $i$.
    The target backbone's own solo LoRA is therefore excluded.
    With six included components, the LOMO sum has rank at most
    $(K-1)r=96$ per module.
\end{itemize}

After the solo LoRAs have been trained, constructing Exact sum and
LOMO sum requires no additional optimization, validation data for
combination, or backbone-specific combination weights.

\noindent\textbf{Evaluation.}
We evaluate all methods using top-1 classification accuracy (\%).
For the cross-backbone response analysis, we report the accuracy gain
over Backbone.
Each experiment is conducted with three random seeds,
and we report the mean and standard deviation over the three runs. The pretrained backbones are evaluated without training and are therefore reported without a standard deviation.

\newcommand{\ms}[2]{%
  \shortstack{$#1$\\[-1pt]\textcolor{gray}{\scriptsize $\pm$#2}}%
}

\begin{table}[tbp]
\centering
\small
\caption{
Top-1 accuracy (\%) averaged over seven backbones.
Average denotes the unweighted mean over the eight datasets.
The top and bottom values denote the mean and standard deviation,
respectively, over three random seeds.
}
\label{tab:main}
\begin{tabular}{lccccccccc}
\toprule
Method
& Cars
& DTD
& EuroSAT
& GTSRB
& MNIST
& RESISC
& SUN397
& SVHN
& Average\\
\midrule
Backbone
& 81.82
& 55.13
& 54.17
& 45.12
& 61.21
& 67.63
& 69.10
& 44.61
& 59.85\\

Solo
& $\ms{85.14}{0.06}$
& $\ms{78.54}{0.11}$
& $\ms{98.55}{0.04}$
& $\ms{96.22}{0.03}$
& $\ms{99.09}{0.01}$
& $\ms{91.23}{0.04}$
& $\ms{79.30}{0.02}$
& $\ms{94.67}{0.02}$ 
& $\ms{90.34}{0.02}$\\

Joint shared
& $\ms{84.63}{0.08}$
& $\ms{76.68}{0.02}$
& $\ms{98.09}{0.03}$
& $\ms{94.63}{0.04}$
& $\ms{98.74}{0.04}$
& $\ms{89.79}{0.07}$
& $\ms{78.25}{0.01}$
& $\ms{93.96}{0.01}$ 
& $\ms{89.34}{0.02}$\\

Exact sum
& $\ms{85.03}{0.13}$
& $\ms{77.89}{0.14}$
& $\ms{98.49}{0.03}$
& $\ms{95.91}{0.03}$
& $\ms{99.03}{0.01}$
& $\ms{91.03}{0.03}$
& $\ms{78.54}{0.02}$
& $\ms{94.55}{0.05}$ 
& $\ms{90.06}{0.03}$\\

LOMO sum
& $\ms{81.72}{0.03}$
& $\ms{55.07}{0.09}$
& $\ms{54.60}{0.26}$
& $\ms{45.27}{0.21}$
& $\ms{61.33}{0.33}$
& $\ms{67.63}{0.06}$
& $\ms{68.53}{0.05}$
& $\ms{45.10}{0.13}$ 
& $\ms{59.91}{0.13}$\\

\bottomrule
\end{tabular}
\end{table}

\subsection{Main Results}

Table~\ref{tab:main} compares Backbone, Solo, Joint shared,
Exact sum, and LOMO sum.
Based on the mean accuracy over three seeds, Exact sum improves
all 56 backbone--task pairs over Backbone.
Its average accuracy is 0.28 percentage points below Solo.
In contrast, LOMO sum remains close to Backbone.
Together, these results support the hypothesis that each backbone
responds mainly to its own component in the superposed update.

Joint shared provides a jointly optimized rank-16 reference.
Once the seven solo LoRAs have been trained, Exact sum reaches
90.06\% accuracy, compared with 89.34\% for Joint shared, and its
construction requires no further access to the training data.
However, the comparison is not rank-matched: Exact sum has rank at
most 112 per module, whereas Joint shared has rank 16.
This comparison therefore concerns functional performance rather
than adapter size or parameter efficiency.
The two approaches also differ when a new backbone is added.
Joint shared requires additional joint optimization involving the
expanded backbone set.
By contrast, Exact sum requires training only a solo LoRA for the new
backbone and then adding its update to the stored superposition.
The previously trained solo LoRAs need not be stored or retrained.

\subsection{Geometric Relation of Solo LoRAs}

To examine the geometric basis of non-interference, we measure the
cosine similarity between solo LoRAs trained for different backbones.
Using all corresponding module updates, we define
\begin{equation}
    \operatorname{cos}
    (\bm{\tau}_{i,t},\bm{\tau}_{m,t})
    =
    \frac{
        \bm{\tau}_{i,t}^{\top}\bm{\tau}_{m,t}
    }{
        \|\bm{\tau}_{i,t}\|_2
        \|\bm{\tau}_{m,t}\|_2
    }.
    \label{eq:lora-cosine}
\end{equation}

\begin{figure}[!t]
    \centering
    \includegraphics[
        width=0.7\columnwidth
    ]{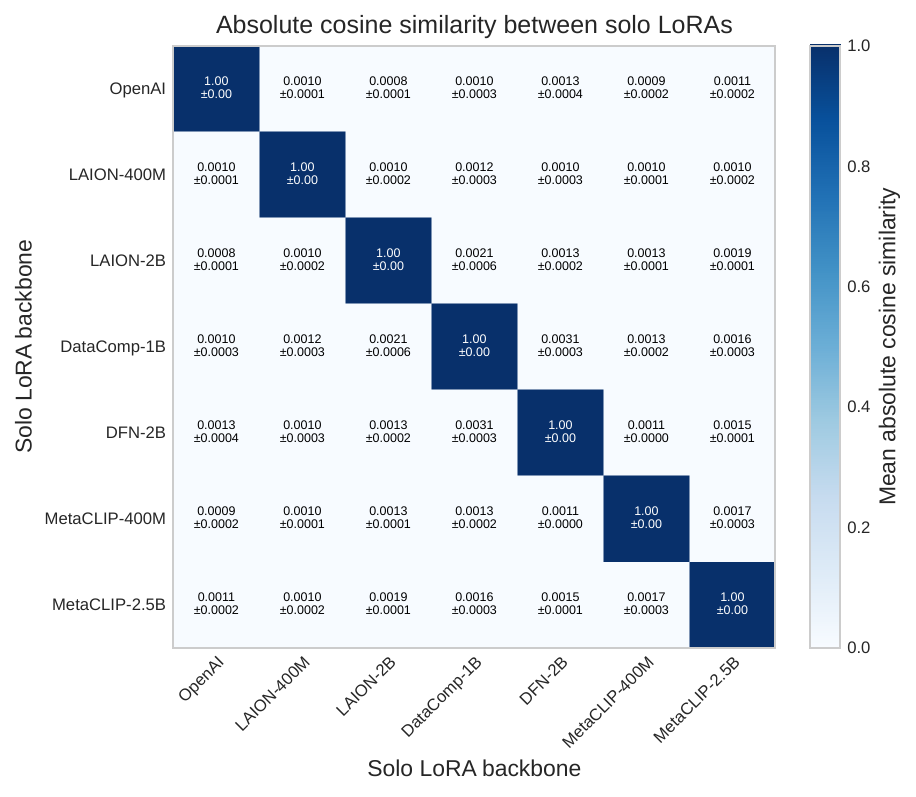}
    \caption{
    Absolute cosine similarity between backbone-specific solo LoRAs,
    averaged over eight tasks.
    The top and bottom values denote the mean and standard deviation,
respectively, over three random seeds.
    }
    \label{fig:lora-cosine}
\end{figure}

Figure~\ref{fig:lora-cosine} shows the absolute cosine similarity
averaged over the eight tasks.
Small off-diagonal values indicate that solo LoRAs trained for
different backbones occupy nearly orthogonal directions in parameter
space.
The mean absolute cosine similarity across different backbones is
$1.34 \times 10^{-3}$.
For reference, the expected absolute cosine similarity between two
independent isotropic Gaussian vectors in $\mathbb{R}^{D}$ is
approximately $\sqrt{2/(\pi D)}$, which is $2.12 \times 10^{-4}$
for $D=24\times768^2$.
In comparison, the mean absolute cosine similarity between solo
LoRAs trained on the same backbone and task with different random
seeds is $0.424$.
Thus, the cross-backbone cosine similarities are substantially smaller than the within-backbone, 
cross-seed similarity. 
However, low cosine similarity alone does not establish functional independence.

\subsection{Pairwise Cross-Backbone Response}

The aggregate LOMO result tests whether the collection of
other-backbone components affects a target backbone.
We further examine the response to each individual solo LoRA.
For every pair of backbones $(i,m)$, we apply
$\bm{\tau}_{m,t}$ to backbone $i$ and compute
\begin{equation}
    a_{i,t}(\bm{\tau}_{m,t})
    -
    a_{i,t}(\bm{0}).
    \label{eq:pairwise-gain}
\end{equation}

\begin{figure}[!t]
    \centering
    \includegraphics[
        width=0.7\columnwidth
    ]{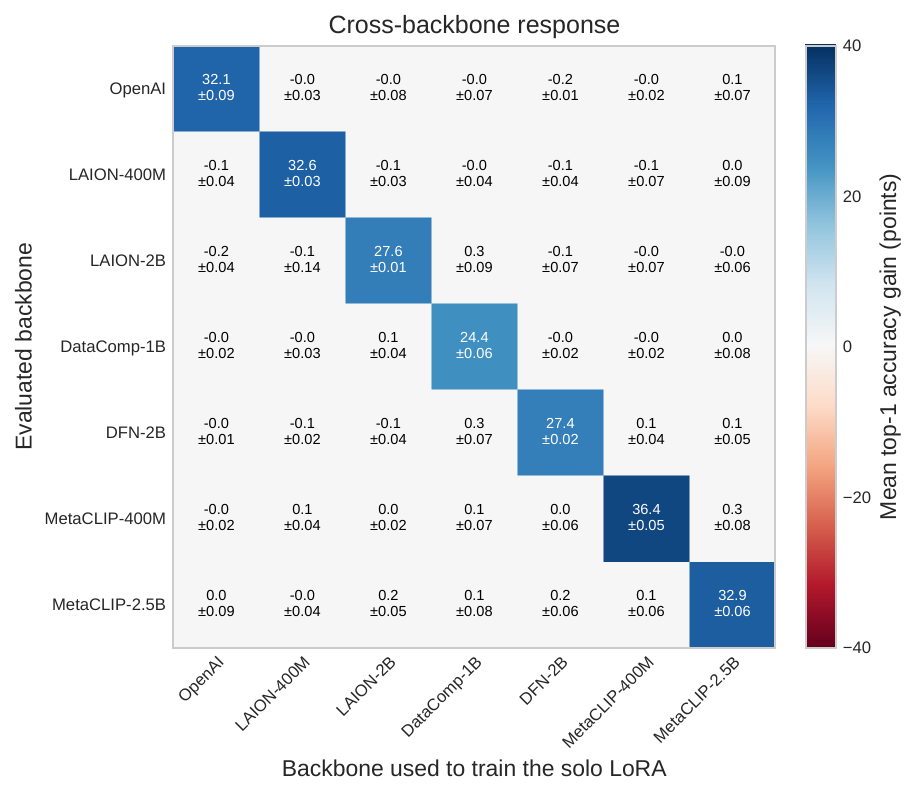}
    \caption{
    Cross-backbone response of individual solo LoRAs.
    Each cell shows the top-1 accuracy gain in percentage points,
    averaged over eight tasks, when the solo LoRA trained for the
    column backbone is applied to the row backbone.
    The top and bottom values denote the mean and standard deviation, respectively, over three random seeds.
    }
    \label{fig:component-response}
\end{figure}

Figure~\ref{fig:component-response} reports this accuracy gain
averaged over the eight tasks.
A diagonal response indicates that a solo LoRA mainly affects the
backbone on which it was trained, while small off-diagonal responses
support pairwise non-interference.

\section{Discussion and Limitations}

The results suggest that backbone-specific LoRAs can coexist in a
shared update because each frozen backbone responds primarily to its
own component.
However, near-orthogonality of the LoRA updates alone does not explain
this behavior.
Joint shared achieves competitive performance with one common update across all seven backbones. This shows that an update can be useful across multiple backbones. The small off-diagonal responses in Fig.~\ref{fig:component-response} may therefore reflect both backbone-specific sensitivity and the separate training of the solo LoRAs.
The key unresolved question is how the geometry of parameter updates
interacts with the backbone-specific directions to which each model is
functionally sensitive.
Our implicit-projection interpretation provides a useful description
of the observed behavior, but the effective subspace induced by each
backbone is not explicitly characterized.
Furthermore, it remains unclear whether the observed near-orthogonality is a general property of independently trained LoRAs or depends on the particular backbones, tasks, and training conditions.
Understanding this relationship would provide a more principled
explanation of when LoRA superposition succeeds or fails.

Another fundamental question is how non-interference scales with the
number and diversity of pretrained backbones.
Small pairwise responses do not necessarily imply that the sum of many
foreign components remains negligible, because weak interactions may
accumulate as the backbone set grows.
The present study also considers backbones with the same ViT-B/16
architecture, where corresponding LoRA modules are directly
compatible.
Extending the superposition principle beyond this setting would
require understanding whether analogous backbone-specific effective
subspaces can be identified across different architectures, rather
than merely aligning their parameters
~\cite{farhadzadeh2025lorax,rinaldi2025transfusion,
kawamoto2025orthogonal}.

The trained solo LoRAs can be combined into one shared adapter without additional optimization. The same adapter can be distributed and applied unchanged to all contributing backbones, removing the need to select a backbone-specific adapter. However, the rank of the superposed update can grow with the number of backbones, with an upper bound of 112 in our setting. Future work could explore compression to rank 16. In factorized form, the compressed update would have the same storage cost as a single solo LoRA. Whether this compression preserves adaptation performance across all contributing backbones remains to be evaluated.

\section{Conclusion}

We study whether backbone-specific solo LoRAs for the same task can
be summed and applied unchanged to a closed set of pretrained models.
The results support LoRA superposition: each frozen backbone responds
mainly to its own component in the summed update.
The near-orthogonal geometry of solo LoRAs, pairwise cross-backbone
responses, and target-excluded results together suggest that
backbone-specific adaptations can coexist in a shared parameter
update without explicit routing or cross-model alignment.

This finding provides a different perspective on the backbone
dependence of LoRA.
Rather than treating backbone-specific adapters as mutually
incompatible objects, LoRA superposition suggests that their effects
can remain functionally separable after addition, with the frozen
backbone itself acting as an implicit selector.
This opens a route toward modular reuse of adapter libraries across
multiple pretrained models without joint retraining.

A central question for future research is why this implicit selection
emerges from the interaction between LoRA update geometry and
backbone-specific functional sensitivity.
Understanding this mechanism may clarify when non-interference holds,
how it scales with the number and diversity of backbones, and whether
an analogous superposition principle extends beyond models with
identical architectures.

\section*{Acknowledgments}
This work was supported by JSPS KAKENHI Grant Number JP26K02967, and the Support Center for Advanced Telecommunications Technology Research (SCAT).

\begingroup
\small
\bibliographystyle{splncs04}
\bibliography{references}
\endgroup
\end{document}